%% file: main.tex
\documentclass[sigconf,nonacm]{acmart}
\renewcommand\footnotetextcopyrightpermission[1]{}
\AtBeginDocument{%
  }

\usepackage{algorithm}
\usepackage{algorithmic}
\usepackage{graphicx}
\usepackage{textcomp}
\usepackage{xcolor}
\usepackage{setspace}
\usepackage{subcaption}
\usepackage{multirow}
\usepackage{array}
\begin{document}

\newcommand{\Brand}{UltraSketchLLM}

\title{UltraSketchLLM: Sub-1-Bit LLM Compression via Sketch and Hardware-Friendly Operators}

\newcommand{\affmark}[1]{\textsuperscript{#1}}

\author{Sunan Zou\affmark{1, 3}, Xueting Sun\affmark{2, 3}, Ziyun Zhang\affmark{1, 3}, and Guojie Luo\affmark{1, 3}}
\affiliation{%
    \textsuperscript{1}National Key Laboratory for Multimedia Information Processing, School of Computer Science, Peking University
    \country{China}
}
 
\affiliation{%
  \textsuperscript{2}School of Electronic Engineering and Computer Science, Peking University
  \country{China}
}
  
\affiliation{%
  \textsuperscript{3}Center for Energy-efficient Computing and Applications, Peking University
  \country{China}
}
  
\email{zousunan@pku.edu.cn, {sunxueting, ziyunzhang}@stu.pku.edu.cn, gluo@pku.edu.cn}

\begin{abstract}
Large language models (LLMs) require larger GPU memory size these days, necessitating efficient and extreme weight compression methods.
Existing compression methods are either theoretically limited by 1 bit per weight or face severe performance degradation and inefficiency.
To deploy LLMs in resource-constrained scenarios, we introduce \Brand{}, compressing LLMs with data sketch.
It reduces peak GPU memory footprint with a high compression rate down to 0.5 bit per weight.
Combined with hardware-friendly implementation, \Brand{} keeps tolerable performance degradation and extremely low latency overhead with $14.9\times$ speedup compared to naive sketch solution.
\footnote{Accepted by the 63rd ACM/IEEE The Chips to Systems Conference (DAC 2026)}
\end{abstract}



\gdef\authors{Sunan Zou, Xueting Sun, Ziyun Zhang, and Guojie Luo}
\renewcommand{\shortauthors}{Sunan Zou, Xueting Sun, Ziyun Zhang, and Guojie Luo}
\maketitle

\input{1-Introduction}
\input{2-Background}
\input{3-Method}
\input{4-Experiment}

\input{5-Conclusion}




\begin{acks}
This work was partially supported by a grant from Infinigence AI. The authors also thank Dr. Shunlin Zeng and Dr. Wenheng Ma of Infinigence AI for their valuable insight.
\end{acks}

\bibliographystyle{ACM-Reference-Format}
\bibliography{reference}


\end{document}

%% file: 1-Introduction.tex
\section{Introduction}
\begin{figure}
\centering
\begin{subfigure}[b]{0.33\linewidth}
\centering
\includegraphics[width=1\columnwidth]{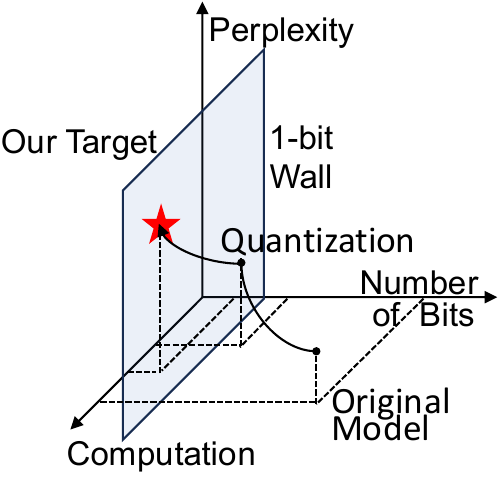}

\caption{}
\label{fig:pareto}
\end{subfigure}
\begin{subfigure}[b]{0.6\linewidth}
\centering
\includegraphics[width=1\columnwidth]{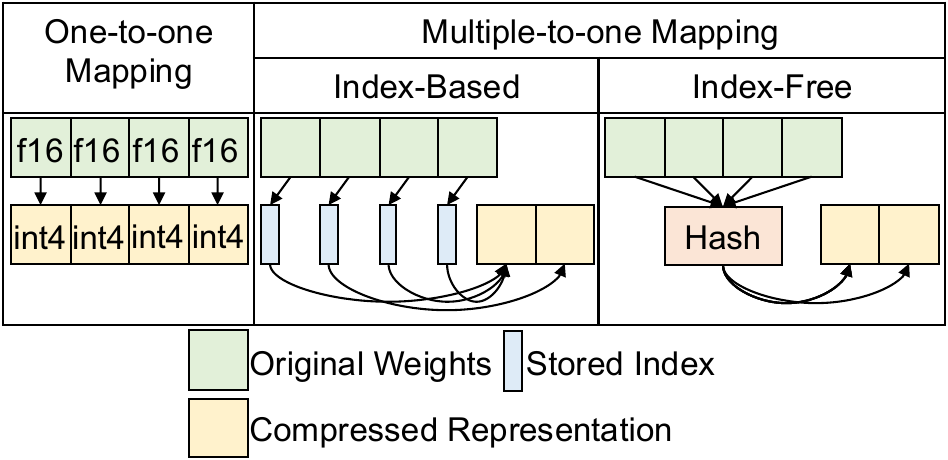}

\caption{}
\label{fig:classification}
\end{subfigure}
\caption{(a) \Brand{} target; (b) Classification of model compression methods.}
\label{fig:intro}
\vspace{-4em}
\end{figure}

The rapid advancement of large language models (LLMs) has unlocked remarkable capabilities in natural language processing, yet their ever-growing scale presents significant hurdles for deployment in resource-constrained environments.
As the need for privacy-preserving, low-latency, and cost-effective inference intensifies, compressing LLMs to operate within tight memory budgets has emerged as a pressing research challenge.
\Brand{} targets compressing weight to the extreme while maintaining tolerable performance and extremely low latency overhead, as in Figure~\ref{fig:pareto}.

Existing model weight compression methods can be classified into one-to-one mapping (OTO), indexed multiple-to-one mapping, and index-free multiple-to-one mapping (MTO), as in Figure~\ref{fig:classification}.
OTO methods like quantization~\cite{lin2024awq} represent each weight independently.
MTO methods group weights by either an index-free method with hash function~\cite{chen2015compressing, saedi2024ss1} or a mapping index~\cite{vptq} for each weight.
Then, grouped weights share one or several representations.
Both OTO and indexed MTO compression are limited to 1 bit per weight, while index-free MTO tends to induce large errors.
Furthermore, the heterogeneity induced by compression, such as hybrid-bit and random address access, severely harms inference efficiency on memory-constrained devices.
The trade-off between compression ratio and computation efficiency is hard to balance.

To tackle these challenges, we introduce \Brand{}\footnote{Code available at https://github.com/suenar/UltraSketchLLM}, an index-free MTO compression scheme leveraging data sketch with an efficient hardware implementation.
Sketch methods approximate data with sub-linear space~\cite{cormode2005improved}, under theoretically bounded estimation error.
They leverage scenario-specific select (data mapping) and update (conflict solving) functions to construct sketch states representing original data.
We propose the underestimating AbsMaxMin sketch with importance-aware sketch space allocation based on our comprehensive profiling results.
However, a sub-1 bit per weight scheme still induces noteworthy error, and \Brand{} adopts compression-aware fine-tuning to mitigate it.
Unlike previous MTO works ~\cite{desai2022efficient, zhang2024spallm} using gradient aggregation, which limits the degree of freedom, \Brand{} leverages a straight-through estimator (STE) with a multi-row AbsMaxMin sketch design.
Such a gradient estimation scheme for discrete sketch mapping enhances the expressiveness of the compressed model.

Sketch disperses weights to sketch states via a random hash to lower global estimation error, which faces a severe computation efficiency issue.
Unlike previous works that increase hash granularity~\cite{desai2022efficient} to reduce overhead at the cost of degree of freedom, \Brand{} maps sketch operators to matrix operations with a tailored kernel design and caching scheme.
Both compression and decompression can be completed within one redefined matrix multiplication of the original weight (compression) or sketch states (decompression) with a mapping matrix.
Such a scheme significantly reduces inference overhead by removing memory access irregularity and incurs almost no storage overhead by tuning the mapping matrix size.

To summarize, our key contributions are:
1) \textbf{Sketch-based LLM compression}.
AbsMaxMin sketch with importance-aware space allocation reduces relative error and preserves salient weights, allowing extreme compression.
2) \textbf{Hardware-friendly compression operator}.
Tailored matrix operation kernels avoid heterogeneity in sketch operations, enhancing computation efficiency.
3) \textbf{Ultra-low bit compression and low overhead}.
Experiments validate that \Brand{} achieves down to 0.5 bit per weight compression with tolerable performance and negligible overhead.

%% file: 2-Background.tex
\section{Background}

\subsection{Compression for LLM}
\label{sub:weight_feature}

\begin{figure}[t]
\centering
\includegraphics[width=0.7\columnwidth]{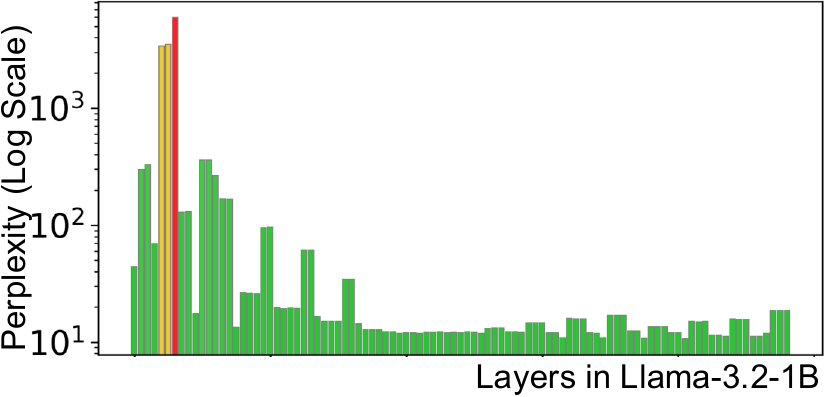}

\caption{Model perplexity after layer-wise compression.}
\label{fig:importance_profiling}
\vspace{-1em}
\end{figure}

Transformer blocks~\cite{vaswani2017attention}, a key component in large language models (LLMs), rely on two primary networks: self-attention (Attn) and feed-forward networks (FFN). 
They are defined by:
\begin{equation}
    \text{Attn}(X) = \text{Softmax}\left( \frac{XW^Q (XW^K)^\top}{\sqrt{d_k}} \right) XW^V \cdot W^O
\end{equation}
\begin{equation}
    \text{FFN}(X) = \left( \text{Activation}(XW^{Gate}) \odot XW^{Up} \right) W^{Down}\text{,}
\end{equation}
where $X$ is the input vector and $W^{\{Q,K,V,...\}}$ are projection matrices, constituting most of the trainable parameters in LLMs.
Compressing these weights is essential to reduce the memory footprint.
Figure~\ref{fig:classification} classifies compression methods into OTO, indexed, and index-free MTO, mapping weights into light representations.

OTO methods map each weight to an exclusive representation, like scalar quantization, representing each weight with a low-bit scalar.
They exploit weight distribution~\cite{huang2024billm}, importance~\cite{yao2021hawq}, and relevance to activation~\cite{xiao2023smoothquant} to reduce performance degradation.
For MTO, a group of weights shares representations, enabling sub-1-bit compression.
Existing index-free MTO~\cite{desai2022efficient, chen2015compressing} can increase errors and therefore concentrate on the pre-training stage, while indexed MTO~\cite{zhang2024spallm, vptq} introduces memory overhead from mapping tables.
They leverage gradient aggregation for compression-aware training, constraining the degree of freedom.
How to fully explore the importance discrepancy among weights for MTO is also untouched.

We profiled the sketch influence on LLM to guide a fitting design.
Different layers of the Llama-3.2-1B model~\cite{grattafiori2024llama} were separately compressed to $\frac{1}{8}$ size using a naive sketch algorithm~\cite{cormode2005improved}.
Figure~\ref{fig:importance_profiling} shows that compression increases perplexity to varying degrees across layers, ranging from $10.9$ to $5950.5$ , compared to the original perplexity of $10.3$.
In addition, the relative error induced by compression techniques strongly influences the model's performance~\cite{gong2024makes}.
Sketch methods preserve more weights untouched (15.29\%) than quantization (3.95\%), while larger relative and sign error result in greater model quality degradation.
Based on the profiling results, \Brand{} uses an underestimating AbsMaxMin sketch and importance-aware space allocation strategy with straight-through estimator-based fine-tuning to squeeze out every bit of representation capability.

Though exploring heterogeneity in LLM weights is a must for a high compression rate like mixed-precision quantization~\cite{zhang2024mixpe}, such a scheme introduces significant computational inefficiencies due to type conversion~\cite{zhang2025hack} and irregular memory access patterns~\cite{zhao2024atom}, especially for MTO methods with random weight mapping.
These inefficiencies necessitate algorithm-system co-design approaches.
Unlike previous works designing specific architectures, which might be impractical for common users, \Brand{} solves the problem at the algorithm level, mapping operators to matrix operations friendly for widely used modern GPUs.

\subsection{Data Sketch Techniques}
Sketch is a sub-linear memory compression technique that approximates large datasets with controlled error bounds~\cite{cormode2005improved}.
It maintains a sketch state (matrix) $S (M \times N)$ as in Figure~\ref{fig:absmaxmin} and follows a select-update-retrieve process:
\textbf{Select}: map data to specific locations in the sketch matrix, generally by a hash function.
\textbf{Update}: sketch state updating rules based on input data and the current state value to solve data conflicts.
\textbf{Retrieve}: access sketch state values and interpret them based on data keys and retrieval rules.
Select and update rules decide the estimation theoretical error bound and efficiency of computation.
During compression, each datum is processed sequentially by select and update operations in a stream, and decompression follows the select-retrieve process.
The error upper bound of sketch is tightly related to the sketch space assigned, as $E(\text{Error}) \propto Mem(Sketch)$.
A larger sketch state leads to a higher approximation accuracy in theory and vice versa.
The agile error control helps \Brand{} adopt an importance-aware weights compression rate by allocating various sketch matrix sizes.
Applying sketch to LLM weights is orthogonal to scalar quantization, which provides significant compression potential in combination.

Substantial memory reduction makes sketch superior in streaming applications like heavy hitters, cardinality~\cite{zou2024musa}, and frequency distributions~\cite{cao2024learning}.
SketchML~\cite{jiang2018sketchml} and SK-Gradient ~\cite{gui2023sk} also apply sketch in the machine learning domain.
spaLLM~\cite{zhang2024spallm} and Ss1~\cite{saedi2024ss1} apply sketch as a parameter sharing technique, ignoring LLM weight features for specific sketch algorithm design.
Besides, hash operation in sketch are originally designed for streaming data processing with random and irregular memory access patterns sequentially, causing inefficiency in LLM computation flow, which prefers regular memory accesses in batch.
\Brand{} maps sketch operations to equivalent matrix operations, friendly for GPU and other tensor processors, fitting downstream LLM operations.

%% file: 3-Method.tex
\section{Design of \Brand{}}
\label{sec:methods}

\begin{figure}[t]
    \centering
    \includegraphics[width=1\columnwidth]{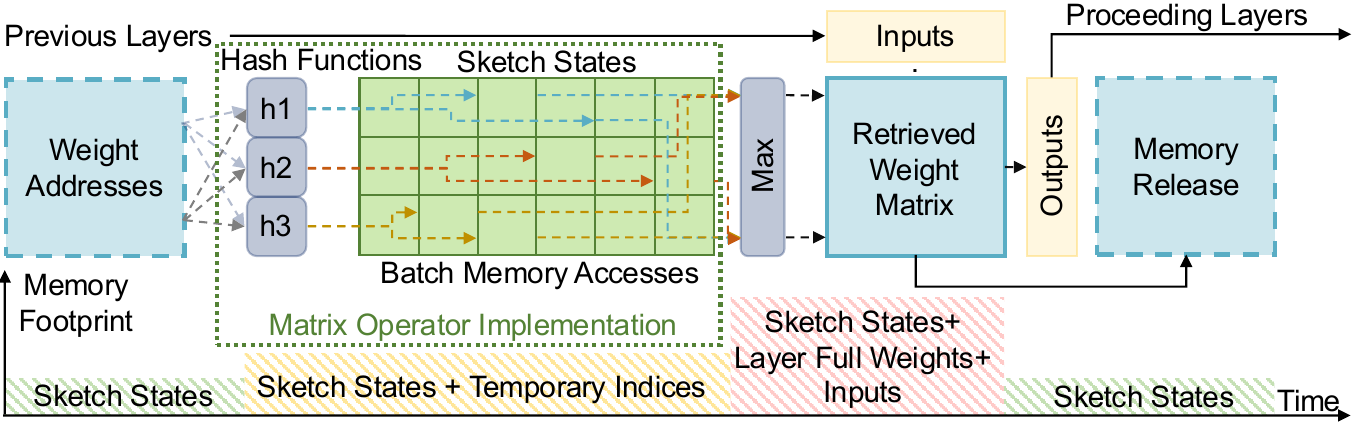}
    \caption{\Brand{} inference procedure.}
    \label{fig:inference_flow}
\end{figure}

\Brand{}, an MTO compression scheme, leverages sketch to represent LLM weights with shared space, which requires much smaller memory than original weight matrices.
Specific sketch algorithm design in Section~\ref{sub:sketch} ensures a high compression rate with tolerable performance degradation, and the hardware implementation provides ultra-low inference overhead.
During LLM computation, sketch states stay permanently in memory, while weights are decompressed dynamically following a layer-by-layer scheme, as in Figure~\ref{fig:inference_flow}.
The forward pass of each layer follows decompression, computation, and memory release stages.
Decompression retrieves weights in batch from sketch states via a single matrix operation.
The computation stage follows the original inference step, using input activation and decompressed weights.
Finally, the memory used in this layer is released, and the next layer computation continues similarly.
Full weights of a particular layer are active only during its calculation, thus the peak memory usage is reduced from $\sum_i^n {Mem(Layer_i)}$ to $\sum_i^n {Mem(Sketch_i)} + Max(Mem(Layer_i))$, where $n$ is the number of layers and $Layer_i$, $Sketch_i$ are $i$-th layer storage by original weight matrix and sketch state, respectively.

\subsection{AbsMaxMin Sketch Design}
\label{sub:sketch}
Sketch compresses a group of unrelated weights to a shared space.
Unlike quantization, whose error is solely controlled and bounded by the original weight, sketch may cause large relative and even sign errors.
Weights with larger absolute values are more resistant to perturbations, and mapping weights with small absolute values to large ones generally causes more severe degradation than the inverse situation~\cite{huang2024billm}.
Therefore, \Brand{} adopts an underestimate AbsMaxMin sketch avoiding such a scenario.
Multiple sketches are used to compress mutually disjoint groups of weights, and \Brand{} explores group size to balance the trade-off between computation efficiency and compression ratio.

\begin{figure}[t]
\centering
\begin{subfigure}[b]{0.55\linewidth}
\centering
\includegraphics[width=1\columnwidth]{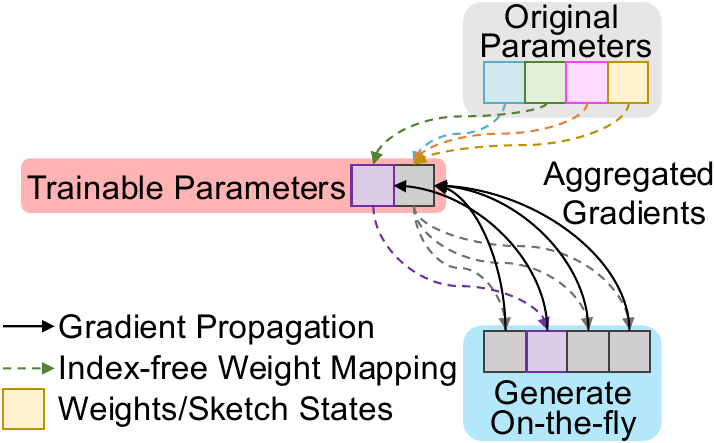}
\vspace{-2em}
\caption{}
\label{fig:aggerated}
\end{subfigure}
\begin{subfigure}[b]{0.55\linewidth}
\centering
\includegraphics[width=1\columnwidth]{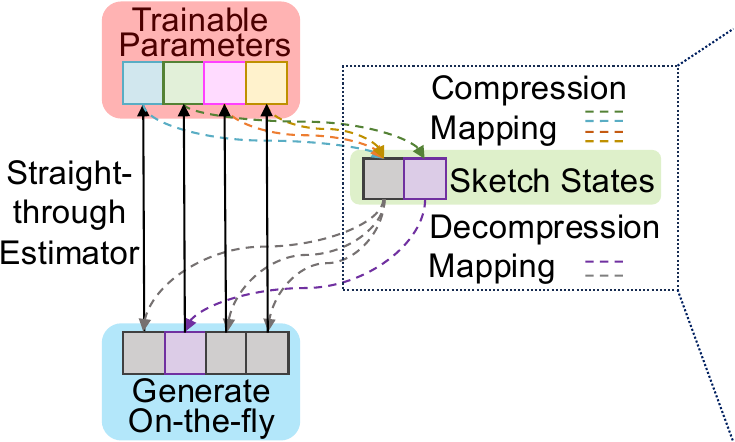}
\caption{}
\label{fig:ste}
\end{subfigure}
\hspace{-0.6em}
\begin{subfigure}[b]{0.4199\linewidth}
\centering
\includegraphics[width=1\columnwidth]{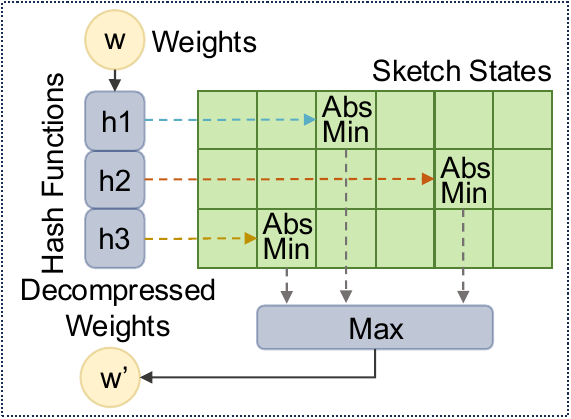}
\caption{}
\label{fig:absmaxmin}
\end{subfigure}
\caption{(a) Aggregated gradient estimation from previous works; (b) Straight-through estimator for fine-tuning in \Brand{}; (c) Multi-row AbsMaxMin sketch design.}
\label{fig:general_flow}
\end{figure}

Sketch design involves three primary components: sketch state $S$, select, and update-retrieve operations.
Sketch states are maintained during compression and approximately represent the original data, whose size can be allocated based on the accuracy needed.
Elements in $S$ store the mapped weights with minimum absolute value and are initialized as $inf$.
Previous index-free MTO mapping for model compression offers only one mapping chance for each weight to a particular shared parameter, as in Figure~\ref{fig:aggerated}.
To introduce multiple choices and create variety during mapping, \Brand{} incorporates a multi-row sketch state design as in Figure~\ref{fig:absmaxmin}.
Each row of $S$ has an independent hash function $H_i$, and each weight is mapped multiple times independently.
Combined with the max operation during retrieval below, such a scheme creates different collision patterns offering more possibilities to group weights with similar values together than a single-row sketch.

During weight compression, the selection function in AbsMaxMin maps each weight to sketch states, which can be formulated by:
\begin{equation}
    I(I_0, I_1, ..., I_n) = H(Addr(w))
\end{equation}
where $w$ is original weight and $H=(H_0, H_1, ..., H_n)$ are independent hash functions.
$Addr(w)$ is the address or offset of the original weight.
AbsMaxMin's update function replaces the sketch state value with the current weight only when its absolute value is smaller than that of the current state, which can be formulated as:
\begin{equation}
    S[i,I_i] = (Abs(S[i, I_i]) > Abs(w))\ ?\ w : S[i, I_i]
\end{equation}
During decompression for inference, we adopt the $Max$ operation to interpret sketch states.
Taking advantage of the mapping pattern variety in a multi-row sketch, the probability of mapping a weight to a close value increases.
This process can be calculated by
\begin{equation}
    w' = Max_i^n(S[i][H_i(Addr(w))])
\end{equation}
Calculations above originally carry on in sequence for each weight, and are replaced by a fused matrix operation for batch processing in real implementation, as explained in Section~\ref{sub:operator_mapping}.

AbsMaxMin, as an underestimate sketch, either preserves the original value or represents it with a smaller absolute value.
Therefore, the error is controlled by the upper bound of the minimum value of sketch states.
The proposed AbsMaxMin sketch ensures that the minimum value exceeds a targeted threshold with high probability, as formulated by the following equation:
\begin{equation}
P(s\ge \Phi^{-1}(1-p^{\frac{1}{n}}))\ge p,
\end{equation}
where $s$ is the sketch state, $\Phi$ is the cumulative distribution function of model weights, $p$ is the target probability, and $n$ is the number of conflicting elements, related to the compression ratio.
The proof establishes a high-probability lower bound for the AbsMaxMin sketch by modeling the number of elements per bucket as binomial and solving for the minimum value threshold in a bucket.

\subsection{Importance-Aware Space Allocation and STE-Based Fine-tuning}

Profiling results in Section~\ref{sub:weight_feature} show that weights vary in their tolerance of perturbation by compression.
As the estimation error is tightly related to the assigned sketch state space, assigning more space to preserve salient weights while applying a high compression rate to others is intuitive.
\Brand{} adopts the expected value of squared activations as an importance metric~\cite{Kawrakow2024imatrix} and assigns sketch space accordingly.
For each row of the weight matrix, the importance value \( I_j \) is given by: $I_j = \mathbb{E}[a_j^2] = \frac{1}{N} \sum_{k=1}^{N} a_{k,j}^2$
where $N$ is the sampled data size, $a_{k,j}$ denotes the $j$-th activation component for the $k$-th sample, and $a_j$ is the $j$-th component of the input activation vector $\mathbf{a}$.
$\mathbb{E}[\cdot]$ represents the expectation over a set of training data.
Notably, row-wise importance can be extended to layer-wise importance by calculating the mean row-wise value across the whole layer.
Given row-wise or layer-wise importance $I_j$, we assign $({I_j} / \sum_i^n I_i) \times Mem(Sketch)$ as the sketch space for each row or layer, where $Mem(Sketch)$ is the total sketch space.

Even with the AbsMaxMin sketch design and importance-aware space allocation, \Brand{} still causes non-negligible relative and sign errors.
Therefore, we apply compression-aware fine-tuning to mitigate model quality degradation, following the routine of previous MTO methods.
Each original weight is first compressed using AbsMaxMin sketch and then retrieved for computation, enabling the fine-tuning process to perceive compression-induced perturbation and update weights to adapt to it.
As hash mapping is discrete, a gradient estimation for this operation is needed.
Previous MTO works use mapped parameters as the only trainable parameters~\cite{zhang2024spallm, desai2022efficient} and estimate the gradient by aggregation, as in Figure~\ref{fig:aggerated}. 
Weights are grouped and mapped only once, limiting the degree of freedom.
Once mapped, values, no matter how large or small, have to converge to a compromised medium value after convergence.

To enhance model expressiveness, we apply a straight-through estimator (STE)~\cite{bengio2013estimating} and only fix the mapping by hash seed, as in Figure~\ref{fig:general_flow}. 
Such a scheme preserves the original trainable parameter size during fine-tuning.
Combined with the multi-row sketch design, \Brand{} offers multiple mapping choices across different rows.
The final mapping choice is dynamically decided by the actual $Min$ and $Max$ result during compression and may change during the fine-tuning process.
As AbsMaxMin sketch considers not only weight mapping but also updated values in sketch states, the fine-tuning process provides another dimension of freedom.

\subsection{Matrix-based Sketch Operator Mapping with Fused Operations}
\label{sub:operator_mapping}
The original sketch implementation is designed for stream processing, hashing each weight to the corresponding sketch states in sequence.
The irregular memory access patterns and conditional branching limit GPU efficiency for LLM compression.
\Brand{} presents a reformulation that maps sketch compression to a redefined matrix multiplication (MM) operation, enabling dramatic performance improvements.
By integrating AbsMaxMin update rules, the hardware-friendly implementation maintains mathematical equivalence while the eliminating heterogeneity induced.

\begin{algorithm}[b]
\caption{Fused Sketch Compression Kernel (CUDA)}
\label{al:kernel}
\begin{algorithmic}[1]
\setstretch{0.85}
\STATE \textbf{Input:} $\mathbf{W}_{\text{g}} \in \mathbb{R}^{M \times G}$, $\mathbf{P} \in \{0,1\}^{G \times K}$
\STATE \textbf{Output:} $\mathbf{S} \in \mathbb{R}^{M \times K}$
\FOR{each thread computing $\mathbf{S}_{i,j}$}
    \STATE $v_{\max} \gets 0$, $\text{abs}_{\min} \gets 0$
    \FOR{$k = 1$ to $G$}
        \STATE $v_{\text{proj}} \gets \mathbf{W}_{\text{g}}[i,k] \cdot \mathbf{P}[k,j]$ \COMMENT{Feature selection}
        \IF{$|v_{\text{proj}}| > \text{abs}_{\max}$}
            \STATE $v_{\max} \gets v_{\text{proj}}$, $\text{abs}_{\max} \gets |v_{\text{proj}}|$ \COMMENT{Sketch update}
        \ENDIF
    \ENDFOR
    \STATE $\mathbf{S}_{i,j} \gets v_{\max}$
\ENDFOR
\end{algorithmic}
\end{algorithm}

Consider a hash function $h: \{1, \ldots, G\} \rightarrow \{1, \ldots, K\}$ that maps weight indices to sketch states, where $G$ and $K$ are group size and sketch size, respectively.
This discrete operation is mathematically equivalent to a projection matrix $\mathbf{P} \in \{0,1\}^{G \times K}$ as:
\begin{equation}
P_{i,j} = [h(i) = j] = \begin{cases}
1 & \text{if index } i \text{ maps to state } j \\
INF & \text{otherwise}
\end{cases}
\label{eq:proj_matrx}
\end{equation}
For the whole compression procedure, we map sketch select function to dot multiplication and update function to $MIN$ reduction, considering absolute value, as in Algorithm~\ref{al:kernel}.
First, we reshape weight matrix $\mathbf{W} \in \mathbb{R}^{d_{\text{in}} \times d_{\text{out}}}$ into groups: $\mathbf{W}_{\text{g}} \in \mathbb{R}^{M \times G}$, where $M = \frac{d_{\text{in}} \cdot d_{\text{out}}}{G}$.
The sketch operation becomes a modified MM as:
\begin{equation}
\mathbf{S}_{i,j} = \underset{k=1}{\overset{G}{\text{min}_{|\cdot|}}} \left( \mathbf{W}_{\text{g}}[i, k] \cdot P_{k,j} \right)
\label{eq:sketch_matrix_form}
\end{equation}
where $\text{min}_{|\cdot|}$ denotes the $MIN$ operation that selects the weight with the smallest absolute value.
Dot multiplication $\cdot$ is also modified to treat $NAN$ as $0$ in case certain weights are $0$.
The compression ratio $\alpha \in (0, 1)$ determines the sketch size $K = \lceil \alpha \cdot G \rceil + 1$ per group, which may vary based on the importance metric.

The redefined MM process above is equivalent to a sequential single-row AbsMaxMin in Section~\ref{sub:sketch}.
Observe that multiplication by $P_{k,j}$ acts as a natural selector: $\mathbf{W}_{\text{g}}[i,k] \cdot P_{k,j}$ equals $\mathbf{W}_{\text{g}}[i,k]$ when $h(k) = j$, and $INF$ otherwise.
The $MIN$ operation then aggregates over all weights mapping to bin $j$, exactly mirroring the hash-based update $\mathbf{S}_{i, h(k)} \gets \max_{|\cdot|}(\mathbf{S}_{i, h(k)}, w_{i,k})$.
Multiplication by $INF$ naturally filters out non-mapped weights without requiring explicit conditional branches when $P_{k,j} = INF$, the product is ignored during minimum absolute value calculation as expected.
Expanding to a multi-row scenario can be carried out by constructing multiple sketch states and MM independently with different projection matrices.
Besides, Algorithm~\ref{al:kernel} fuses the select and update procedure into a single kernel, further enhancing efficiency.

The decompression reconstructs the weight matrix $\hat{\mathbf{W}}$ during inference, implemented by a single general MM operation as:
\begin{equation}
\hat{\mathbf{W}}_{\text{g}} = \mathbf{S} \cdot \mathbf{P}^{\top}
\end{equation}
Since $\mathbf{P}$ is constructed by hash result in Equation~\ref{eq:proj_matrx}, each reconstructed weight retrieves the sketch value from its assigned state and non-mapped sketch states will not be retrieved resulting in $INF$ results: $\hat{\mathbf{W}}_{\text{g}}[i,k] = \mathbf{S}[i, h(k)]$.
The process is equivalent to the lookup operation $\mathbf{S}[i, h(k)]$, but executed as an MM operation.
In summary, both compression and decompression only interact with coalesced memory access patterns, which incur almost no overhead compared to managing sparse index structures in sequence.

\begin{figure}[t]
    \centering
    \includegraphics[width=1\columnwidth]{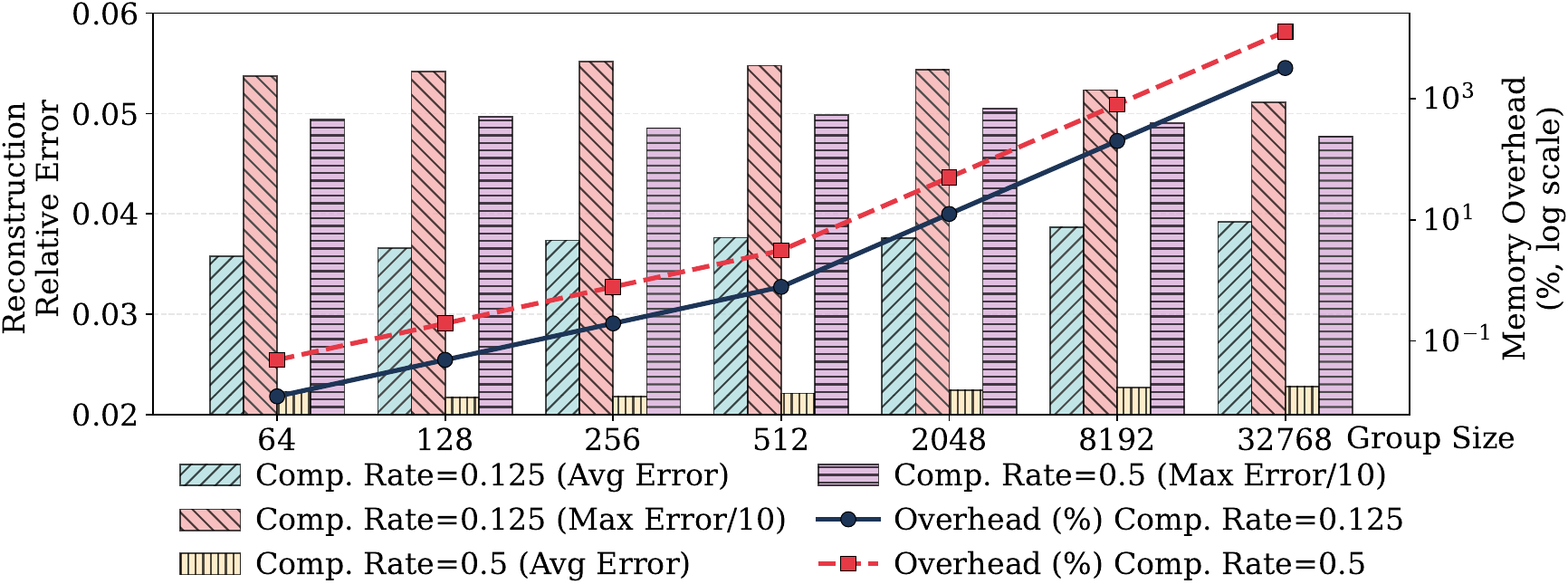}
    \caption{Sketch group size profiling.}
    \label{fig:error_overhead}
\end{figure}

Unlike the original sketch, which solely uses hash for data mapping, matrix-based implementation incorporates a projection matrix $\mathbf{P}$, inducing storage overhead.
However, for the LLM compression task, this overhead is negligible.
Storage of $\mathbf{P}$ is determined by $G \times K$ space, where $G$ is the group size, $K$ is related to the compression ratio, which is constant for a specific setting.
Group size determines both storage overhead and sketch compression effect, as tiny sketch states cannot distribute and hash weights evenly, harming compression quality. 
We profile the balance between overhead and induced error in Figure~\ref{fig:error_overhead}, and select $512$ as the group size, which consumes less than $0.8\%$ storage for Llama-3.2-1B model and will be much smaller as models scale.
Besides, we implement a two-level caching strategy for the projection matrix: each layer maintains an instance cache referencing its projection matrix, while a global cache shares matrices across all layers with identical configurations.
This reduces the amortized overhead from $O(G \cdot K)$ per forward pass to $O(1)$, achieving over $1000\times$ speedup in our experiments compared to regenerating the matrix each time.
The latency overhead during inference is a single general MM operation per layer at a complexity of $O(M \cdot K \cdot G) = O(d_{\text{in}} \cdot d_{\text{out}} \cdot \alpha)$.

The proposed sketch-based compression method is orthogonal to scalar quantization works.
We combine \Brand{} with 4-bit and 8-bit quantization to compress LLM weights to the extreme.
Experiments show this combination induces slightly more perturbation than sketch-based compression alone.

\subsection{Transfer Learning and Parameter Freezing}
While sketch-aware fine-tuning mitigates performance degradation, its training requirement is not always preferable.
\Brand{} introduces a transfer learning approach with parameter freezing, significantly reducing fine-tuning costs.
Analysis of layer-wise tolerance to sketch-induced perturbations in Figure~\ref{fig:importance_profiling} reveals two key findings:
1) Attn layers exhibit greater error resistance than FFN layers.
2) Front-end FFN layers are more vulnerable than back-end FFN layers.
This aligns with findings that FFN layers primarily store task-specific knowledge while Attn layers maintain reasoning capabilities~\cite{geva2020transformer}.
Consequently, Attn layers preserve their logical functions across datasets, whereas FFN layers—especially front-end ones—require dataset-specific tuning.
During transfer learning, we freeze all Attn layers and back-end FFN layers, focusing updates only on critical front-end FFN parameters. 
When adapting a sketch-aware fine-tuned model to a new dataset, this method requires far fewer iterations and trainable parameters.

%% file: 4-Experiment.tex
\section{Experimental Results}
\label{sec:experiments}

\Brand{} is evaluated on Llama3~\cite{grattafiori2024llama} and Qwen3~\cite{yang2025qwen3} families to validate the compression effect, model performance, and inference overhead.
Integration with the HuggingFace framework~\cite{wolf2019huggingface} makes it easy to fine-tune and run LLM models compressed by \Brand{}.
We use importance metrics from~\cite{Kawrakow2024imatrix} for sketch space allocation and combine AbsMaxMin with bitsandbytes~\cite{dettmers2022llmint8} scalar quantization for a high compression rate.
The fine-tuning stage and importance metric calibration utilize the Wikitext~\cite{merity2016pointer} dataset with maximum context length set to $512$.
The transfer learning experiment is conducted on Yelp reviews~\cite{zhang@2015character}.
For inference and compressed model performance evaluation, we test both perplexity on the Wikitext test dataset and downstream tasks, including HellaSwag~\cite{zellers2019hellaswag}, PIQA~\cite{Bisk2020piqa}, and WinoGrande~\cite{ai2:winogrande}.
The fine-tuning process is executed on two NVIDIA A100 GPUs.
During fine-tuning, we set the learning rate in a linear-decay style, starting from $5\times10^{-5}$ and using cross entropy as the loss function.
We compare \Brand{} with a popular quantization method, AQLM~\cite{pmlr-v235-egiazarian24a}~\footnote{As AQLM does not support Qwen3 families, it is removed from part of the experiments.} and an MTO method, VPTQ~\cite{vptq}.

\subsection{LLM Compression Effect and Performance}

\begin{figure}[t]
    \centering
    \includegraphics[width=0.9\columnwidth]{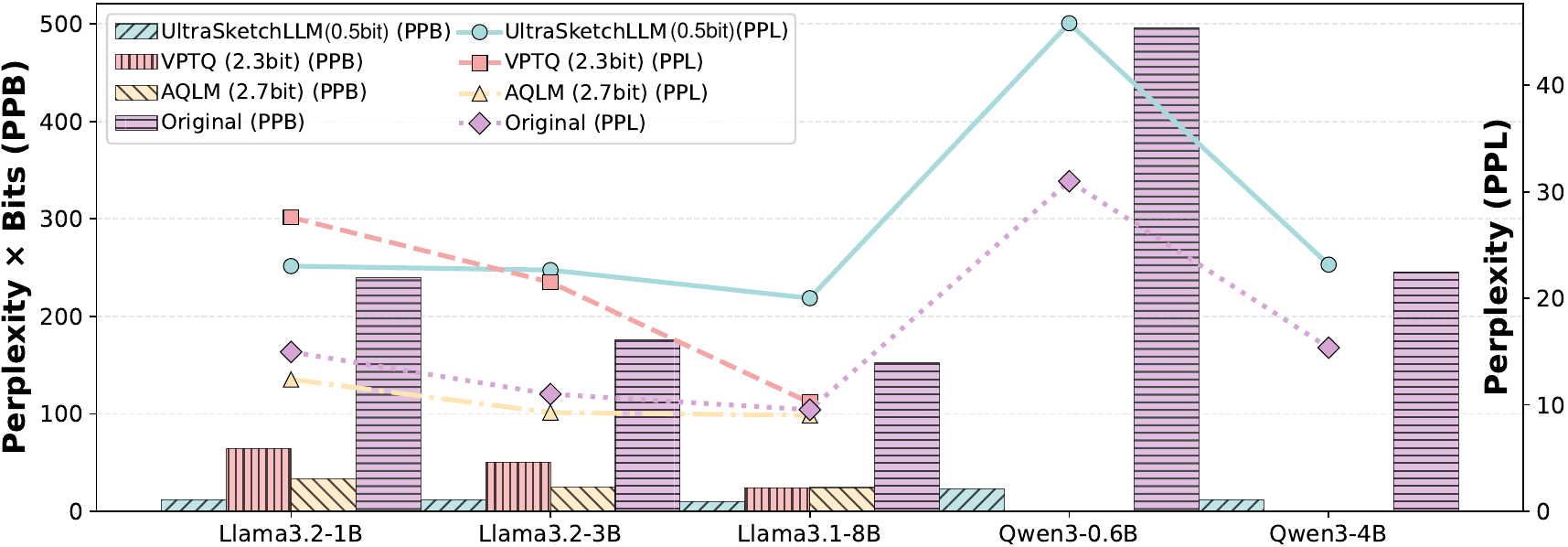}
    \caption{Comparison of PPB and PPL results.}
    \label{fig:ppb_ppl}
\end{figure}

\begin{figure}[t]
    \centering
    \includegraphics[width=0.9\columnwidth]{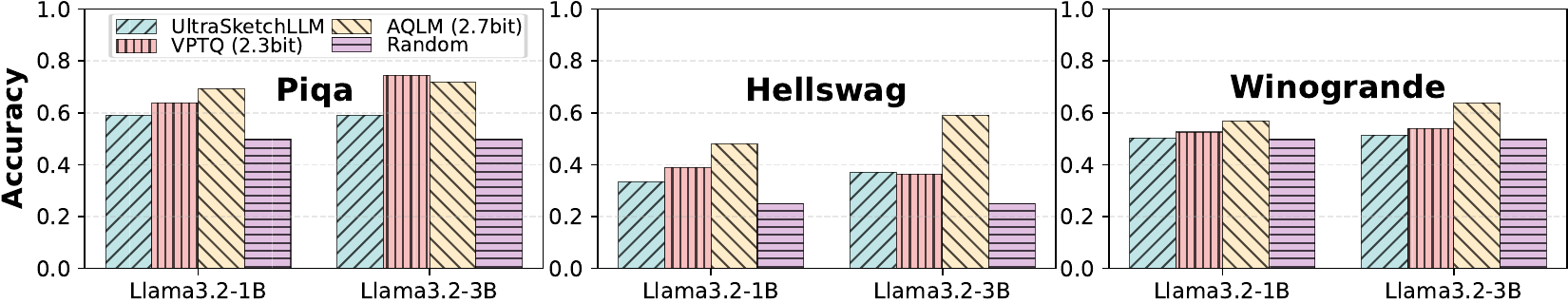}
    \caption{\Brand{} on downstream tasks.}
    \label{fig:downstream}
\end{figure}

\begin{figure}[t]
    \centering
    \includegraphics[width=0.9\columnwidth]{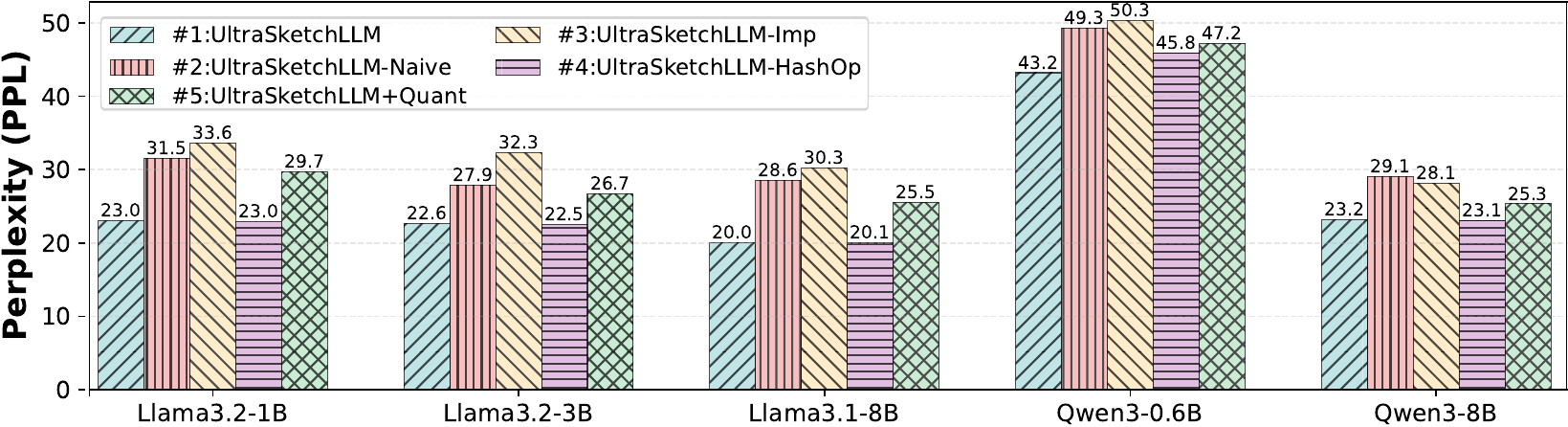}
    \caption{Ablation study shown in perplexity.}
    \label{fig:ablation}
\end{figure}

\Brand{} reaches 0.5 bit per weight compression with comparable model performance when combining $1/8$ sketch compression rate with 4-bit quantization from the original f16 model.
We take perplexity (PPL) and the product of perplexity and average bits (PPB) for original models (f16) as metrics.
Lower perplexity corresponds to better model performance, and smaller PPB means a better balance between higher compression rate and model performance.
As in Figure~\ref{fig:ppb_ppl}, \Brand{} shows much smaller PPB across all settings with up to $26\times$ and $2.43\times$ reduction compared to original and best baseline models, respectively.
The perplexity alone is also comparable, considering \Brand{} achieves a much higher compression rate than baselines.
\Brand{} also shows tolerable performance on downstream tasks as in Figure~\ref{fig:downstream}.
Besides, \Brand{} is orthogonal to quantization methods, and representing sketch states in low-bit types ($\#5$) induces slight quality degradation, as shown in Figure~\ref{fig:ablation}.

We conduct an ablation study to validate the effectiveness of the AbsMaxMin sketch design, matrix-based sketch operation mapping, and importance-aware space allocation scheme.
Setting $\#1$ is the full version of \Brand{}.
Setting $\#2$ applies a naive count-min sketch for compression, which is originally designed for cardinality estimation, simply adding incoming values to existing sketch states and ignoring LLM weight features.
Setting $\#3$ ignores importance variance across LLM weights, showing performance degradation compared to the full version.
\Brand{} also outperforms the naive sketch design significantly.
The matrix-based sketch operation mapping is theoretically equivalent to the original version ($\#4$).
Comparison between $\#1$ and $\#4$ further validates this, showing the same performance under the same random seed.

\subsection{Memory Footprint and Latency Overhead}

\begin{figure}
    \centering
    \includegraphics[width=1\columnwidth]{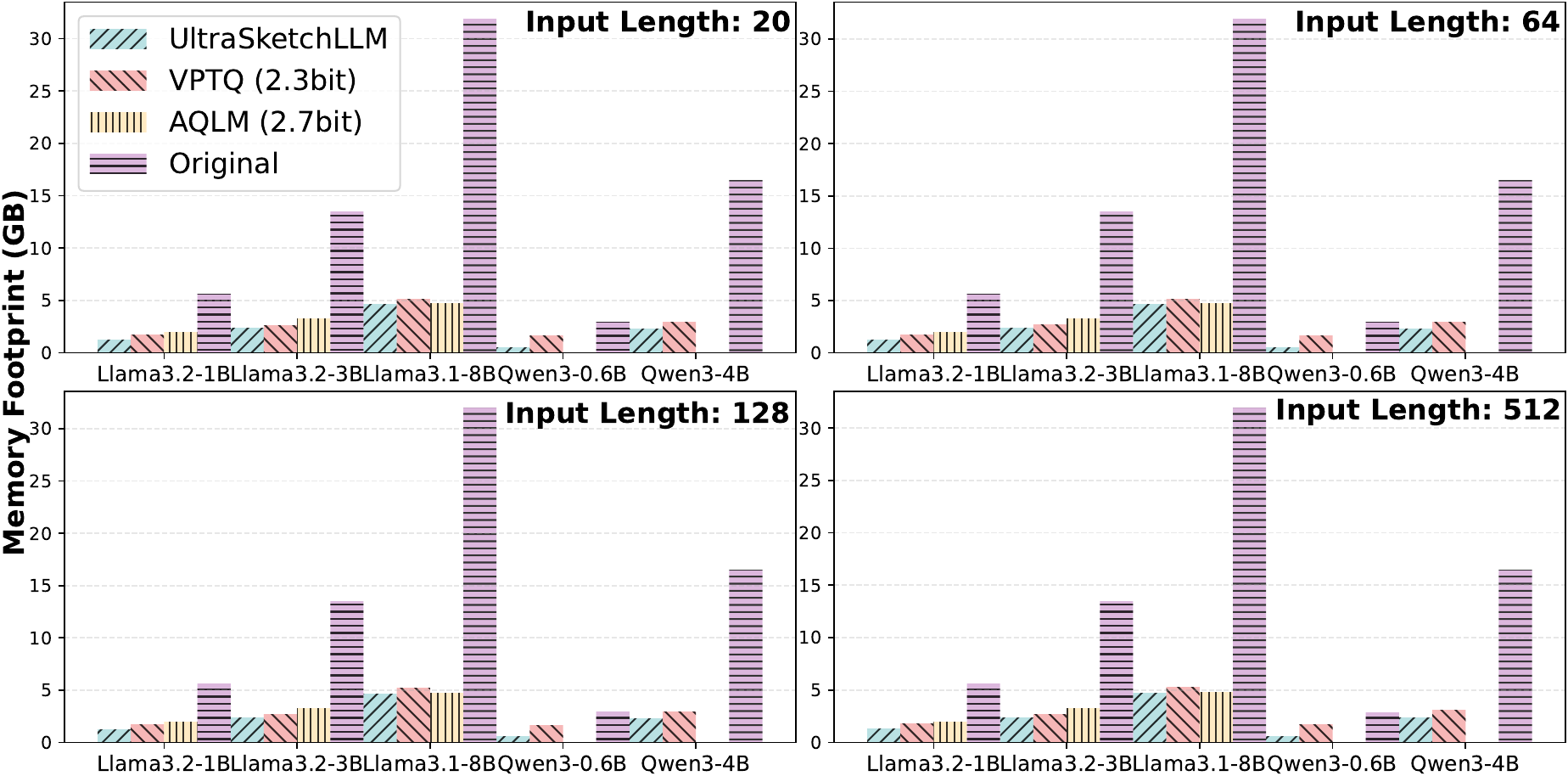}
    \caption{Comparison of memory footprint.}
    \label{fig:memory}
\end{figure}

\begin{figure}
    \centering
    \includegraphics[width=1\columnwidth]{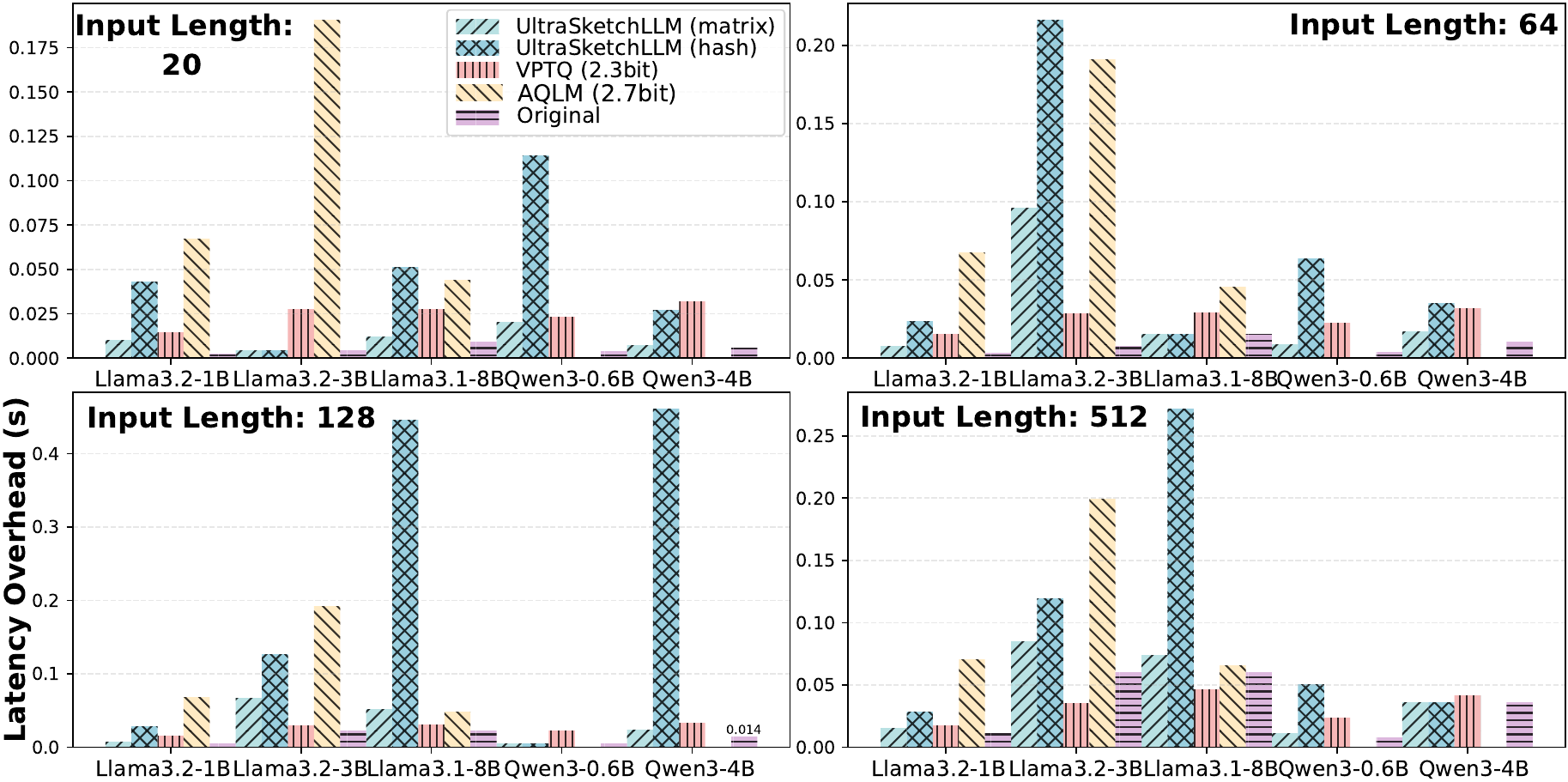}
    \caption{Comparison of first token latency overhead.}
    \label{fig:latency}
\end{figure}

\Brand{} targets reducing the peak memory footprint of LLM inference to an extreme for memory-limited scenarios such as edge devices.
Figure~\ref{fig:memory} and Figure~\ref{fig:latency} showcase the maximum memory footprint and first token latency overhead using \Brand{}, respectively, to validate its actual effect on both memory reduction and latency overhead.
\Brand{} achieves up to $6.93\times$ reduction in memory and outperforms all baselines across all tested settings and models.
Simultaneously, \Brand{} induces less latency overhead than all baselines on all input length scenarios for various models.
The compressed Llama-3.1-8B model can even run on a desktop GPU like NVIDIA GeForce RTX 3080 Ti using only half of its memory.

\begin{figure}
\begin{subfigure}[b]{1\linewidth}
    \includegraphics[width=1\columnwidth]{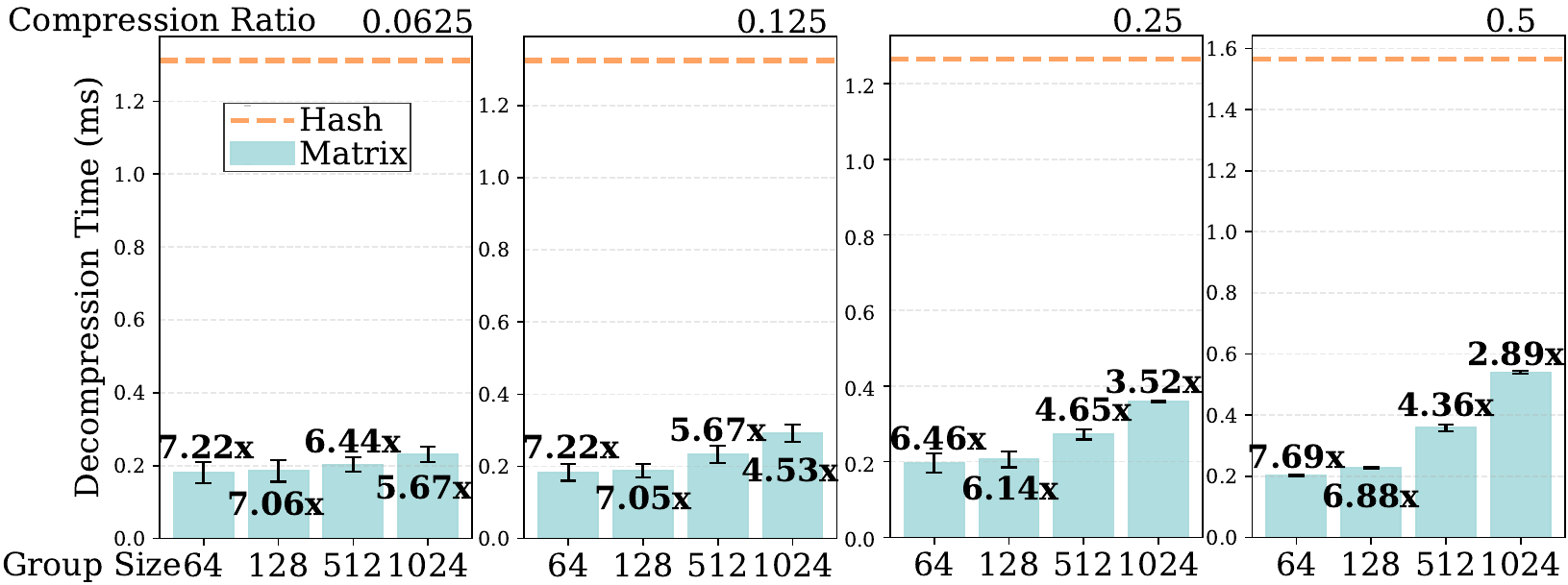}
    \label{fig:speedup_group}
\end{subfigure}
\begin{subfigure}[b]{0.9\linewidth}
    \includegraphics[width=1\columnwidth]{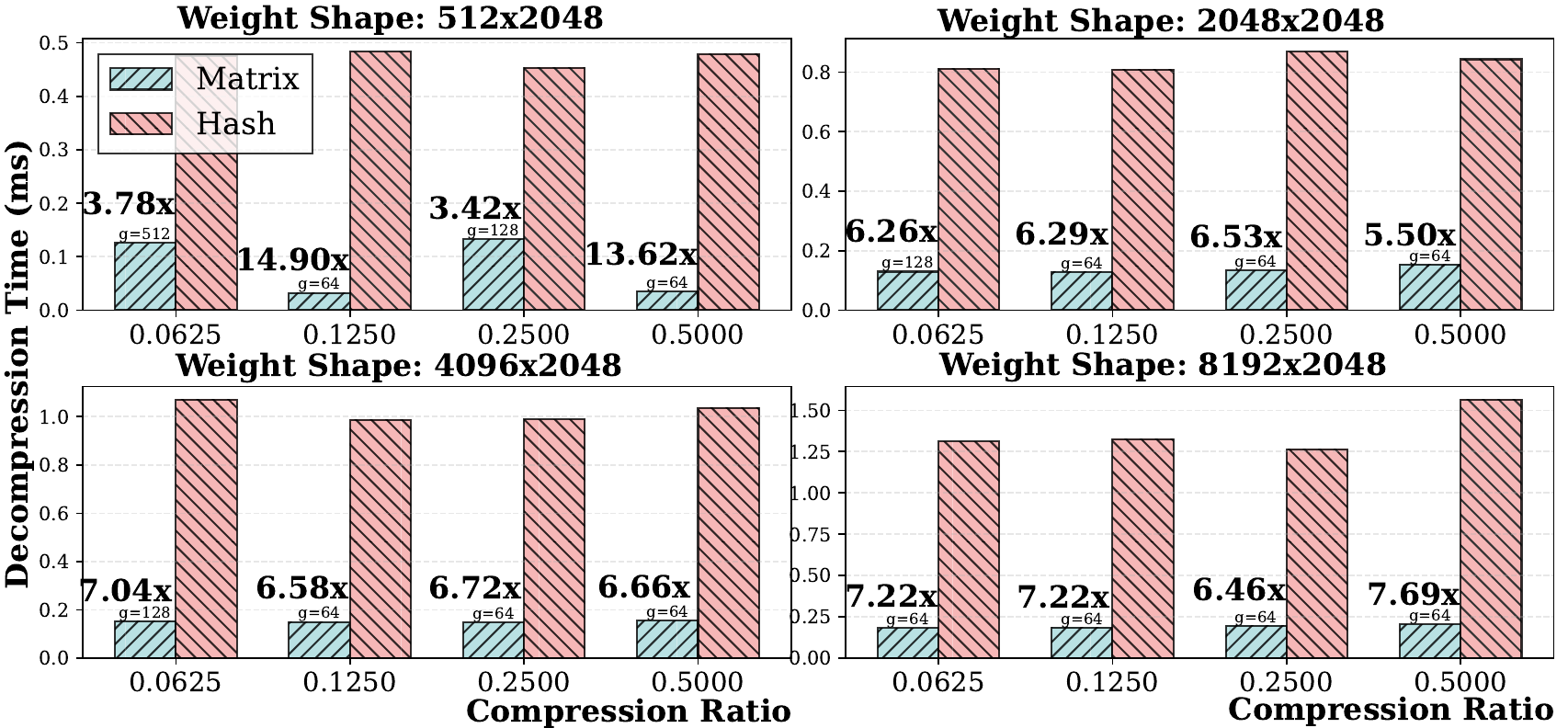}
    \label{fig:speedup_weight}
\end{subfigure}
\caption{Decompression speedups with different group size and weight shapes.}
\label{fig:speedup}
\end{figure}

Matrix-based sketch operation successfully reduces the inference latency overhead to almost negligible compared to the original sketch implementation.
Figure~\ref{fig:speedup} compares the single sketch compression operation latency between matrix operation and hash operation on different weight sizes and group sizes.
Matrix operation outperforms hash operation on all settings, and a maximum $14.9\times$ speedup is achieved.
For the complete inference procedure, matrix-based compression operator mapping transforms intolerable overhead into a negligible one across all generation settings with various input and output lengths, as in Figure~\ref{fig:latency} showing \Brand{} with hash and matrix implementations.
\Brand{} outperforms baselines in inference latency on all tested cases.

\subsection{Transfer Learning and Parameter Freezing}
\label{sub:transfer}

\begin{figure}[t]
    \centering
    \includegraphics[width=0.95\columnwidth]{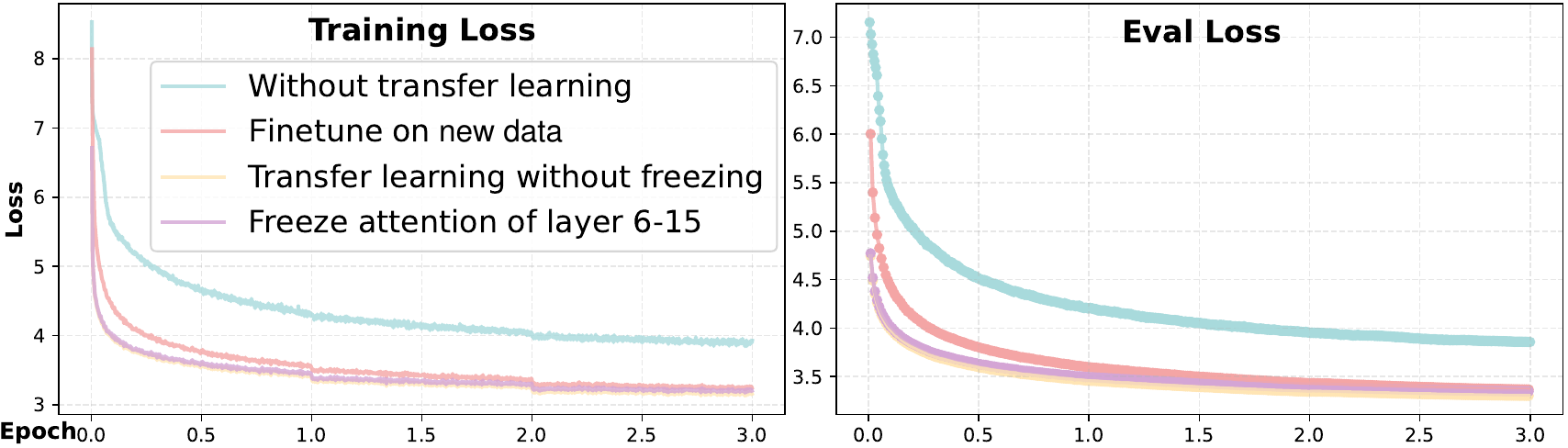}
    \caption{Fine-tuning process with transfer learning and parameter freezing.}
    \label{fig:converging_speed}
    \vspace{-1em}
\end{figure}

To reduce fine-tuning cost for practicability, \Brand{} supports transfer learning to a new dataset from an existing sketch-compressed model with parameter freezing.
Figure~\ref{fig:converging_speed} validates that the proposed transfer learning significantly accelerates convergence.
Besides, freezing Attn layers and back-end FFN layers maintains model quality comparable to full-parameter fine-tuning with less than $4\%$ perplexity increase, as in Figure~\ref{fig:converging_speed}.
This efficiency also helps validate that models previously fine-tuned on source datasets have already adapted to sketch-induced perturbations.

%% file: 5-Conclusion.tex
\section{Conclusion}
In this work, we introduce \Brand{}, an ultra-low bit compression framework that leverages data sketch techniques to deploy LLMs in resource-constrained environments.
Through AbsMaxMin sketch design with importance-aware space allocation and STE-based fine-tuning, we achieve extreme compression rates down to 0.5 bits per weight while maintaining competitive model performance.
Our hardware-friendly operator mapping transforms irregular sketch operations into matrix operations, achieving $14.9\times$ speedup with negligible inference overhead.
Experimental results on Llama3 and Qwen3 families demonstrate substantial memory reduction, up to $6.93\times$ compared to FP16 baselines.
The proposed transfer learning approach further reduces adaptation costs, making \Brand{} a practical solution for edge deployment.

%% file: reference.bib
@inproceedings{gui2023sk,
  title={{Sk-Gradient}: Efficient communication for distributed machine learning with data sketch},
  author={Gui, Jie and Song, Yuchen and Wang, Zezhou and He, Chenhong and Huang, Qun},
  booktitle={IEEE 39th International Conference on Data Engineering (ICDE)},
  pages={2372--2385},
  year={2023},
  _organization={IEEE}
}

@inproceedings{jiang2018sketchml,
  title={{SketchML}: Accelerating distributed machine learning with data sketches},
  author={Jiang, Jiawei and Fu, Fangcheng and Yang, Tong and Cui, Bin},
  booktitle={Proceedings of the 2018 International Conference on Management of Data (SIGMOD)},
  pages={1269--1284},
  year={2018}
}

@inproceedings{chen2015compressing,
  title={Compressing neural networks with the hashing trick},
  author={Chen, Wenlin and Wilson, James and Tyree, Stephen and Weinberger, Kilian and Chen, Yixin},
  booktitle={International Conference on Machine Learning (ICML)},
  pages={2285--2294},
  year={2015},
  _organization={PMLR}
}

@article{zhang2024spallm,
  title={{SpaLLM}: Unified Compressive Adaptation of Large Language Models with Sketching},
  author={Zhang, Tianyi and Su, Junda and Wu, Oscar and Xu, Zhaozhuo and Shrivastava, Anshumali},
  journal={arXiv preprint arXiv:2410.06364},
  year={2024}
}

@inproceedings{zou2024musa,
  title={{MuSA}: Multi-Sketch Accelerator with Hybrid Parallelism and Coalesced Memory Organization},
  author={Zou, Sunan and Shi, Bizhao and Zhang, Ziyun and Luo, Guojie},
  booktitle={IEEE 42nd International Conference on Computer Design (ICCD)},
  pages={36--43},
  year={2024},
  _organization={IEEE}
}

@inproceedings{xiao2023smoothquant,
  title={{SmoothQuant}: Accurate and efficient post-training quantization for large language models},
  author={Xiao, Guangxuan and Lin, Ji and Seznec, Mickael and Wu, Hao and Demouth, Julien and Han, Song},
  booktitle={International Conference on Machine Learning},
  pages={38087--38099},
  year={2023},
  organization={PMLR}
}

@article{cao2024learning,
  title={Learning to sketch: A neural approach to item frequency estimation in streaming data},
  author={Cao, Yukun and Feng, Yuan and Wang, Hairu and Xie, Xike and Zhou, S Kevin},
  journal={IEEE Transactions on Pattern Analysis and Machine Intelligence},
  year={2024},
  publisher={IEEE}
}

@article{vaswani2017attention,
  title={Attention is all you need},
  author={Vaswani, Ashish and Shazeer, Noam and Parmar, Niki and Uszkoreit, Jakob and Jones, Llion and Gomez, Aidan N and Kaiser, {\L}ukasz and Polosukhin, Illia},
  journal={Advances in neural information processing systems},
  volume={30},
  year={2017}
}

@article{desai2022efficient,
  title={Efficient model compression with Random Operation Access Specific Tile (ROAST) hashing},
  author={Desai, Aditya and Zhou, Keren and Shrivastava, Anshumali},
  journal={arXiv preprint arXiv:2207.10702},
  year={2022}
}

@inproceedings{yao2021hawq,
  title={{HAWQ-V3}: Dyadic neural network quantization},
  author={Yao, Zhewei and Dong, Zhen and Zheng, Zhangcheng and Gholami, Amir and Yu, Jiali and Tan, Eric and Wang, Leyuan and Huang, Qijing and Wang, Yida and Mahoney, Michael and others},
  booktitle={International Conference on Machine Learning},
  pages={11875--11886},
  year={2021},
  organization={PMLR}
}

@article{huang2024billm,
  title={{BiLLM}: Pushing the limit of post-training quantization for {LLMs}},
  author={Huang, Wei and Liu, Yangdong and Qin, Haotong and Li, Ying and Zhang, Shiming and Liu, Xianglong and Magno, Michele and Qi, Xiaojuan},
  journal={arXiv preprint arXiv:2402.04291},
  year={2024}
}

@inproceedings{gong2024makes,
  title={What makes quantization for large language model hard? an empirical study from the lens of perturbation},
  author={Gong, Zhuocheng and Liu, Jiahao and Wang, Jingang and Cai, Xunliang and Zhao, Dongyan and Yan, Rui},
  booktitle={Proceedings of the AAAI Conference on Artificial Intelligence},
  volume={38},
  pages={18082--18089},
  year={2024}
}

@article{cormode2005improved,
  title={An improved data stream summary: the count-min sketch and its applications},
  author={Cormode, Graham and Muthukrishnan, Shan},
  journal={Journal of Algorithms},
  volume={55},
  number={1},
  pages={58--75},
  year={2005},
  publisher={Elsevier}
}

@article{bengio2013estimating,
  title={Estimating or propagating gradients through stochastic neurons for conditional computation},
  author={Bengio, Yoshua and L{\'e}onard, Nicholas and Courville, Aaron},
  journal={arXiv preprint arXiv:1308.3432},
  year={2013}
}

@misc{Kawrakow2024imatrix,
  author = {Kawrakow},
  title = {Importance Matrix calculation},
  year = {2024},
  howpublished = {\url{https://github.com/ggml-org/llama.cpp/pull/4861}},
  note = {[GitHub Pull Request \#4861; Access Date: April 15, 2025]},
}

@inproceedings{vptq,
  title={{VPTQ}: Extreme Low-bit Vector Post-Training Quantization for Large Language Models},
  author={Yifei Liu and
          Jicheng Wen and
          Yang Wang and
          Shengyu Ye and
          Li Lyna Zhang and
          Ting Cao and
          Cheng Li and
          Mao Yang},
  booktitle={The 2024 Conference on Empirical Methods in Natural Language Processing},
  year={2024},
}

@article{zhang2024mixpe,
  title={{MixPE}: Quantization and hardware co-design for efficient {LLM} inference},
  author={Zhang, Yu and Wang, Mingzi and Zou, Lancheng and Liu, Wulong and Zhen, Hui-Ling and Yuan, Mingxuan and Yu, Bei},
  journal={arXiv preprint arXiv:2411.16158},
  year={2024}
}

@article{grattafiori2024llama,
  title={The {Llama} 3 herd of models},
  author={Grattafiori, Aaron and Dubey, Abhimanyu and Jauhri, Abhinav and Pandey, Abhinav and Kadian, Abhishek and Al-Dahle, Ahmad and Letman, Aiesha and Mathur, Akhil and Schelten, Alan and Vaughan, Alex and others},
  journal={arXiv preprint arXiv:2407.21783},
  year={2024}
}

@article{merity2016pointer,
  title={Pointer sentinel mixture models},
  author={Merity, Stephen and Xiong, Caiming and Bradbury, James and Socher, Richard},
  journal={5th International Conference on Learning Representations (ICLR 2017)},
  year={2016}
}

@inproceedings{lin2024awq,
  title={{AWQ}: Activation-aware weight quantization for on-device {LLM} compression and acceleration},
  author={Lin, Ji and Tang, Jiaming and Tang, Haotian and Yang, Shang and Chen, Wei-Ming and Wang, Wei-Chen and Xiao, Guangxuan and Dang, Xingyu and Gan, Chuang and Han, Song},
  journal={Proceedings of Machine Learning and Systems (MLSys)},
  volume={6},
  pages={87--100},
  year={2024}
}

@article{wolf2019huggingface,
  title={Huggingface's transformers: State-of-the-art natural language processing},
  author={Wolf, Thomas and Debut, Lysandre and Sanh, Victor and Chaumond, Julien and Delangue, Clement and Moi, Anthony and Cistac, Pierric and Rault, Tim and Louf, R{\'e}mi and Funtowicz, Morgan and others},
  journal={arXiv preprint arXiv:1910.03771},
  year={2019}
}

@inproceedings{dettmers2022llmint8,
  title={LLM.int8(): 8-bit Matrix Multiplication for Transformers at Scale},
  author={Dettmers, Tim and Lewis, Mike and Belkada, Younes and Zettlemoyer, Luke},
  journal={Advances in Neural Information Processing Systems (NeurIPS)},
  year={2022},
}

@article{geva2020transformer,
  title={Transformer feed-forward layers are key-value memories},
  author={Geva, Mor and Schuster, Roei and Berant, Jonathan and Levy, Omer},
  journal={arXiv preprint arXiv:2012.14913},
  year={2020}
}

@inproceedings{zhang@2015character,
  title={Character-level Convolutional Networks for Text Classification},
  author={Xiang, Zhang and Junbo, Zhao and Yann, LeCun},
  booktitle={Advances in Neural Information Processing Systems},
  year={2015}
}

@article{zhao2024atom,
  title={Atom: Low-bit quantization for efficient and accurate {LLM} serving},
  author={Zhao, Yilong and Lin, Chien-Yu and Zhu, Kan and Ye, Zihao and Chen, Lequn and Zheng, Size and Ceze, Luis and Krishnamurthy, Arvind and Chen, Tianqi and Kasikci, Baris},
  journal={Proceedings of Machine Learning and Systems},
  volume={6},
  pages={196--209},
  year={2024}
}

@inproceedings{zhang2025hack,
  title={Hack: Homomorphic acceleration via compression of the key-value cache for disaggregated {LLM} inference},
  author={Zhang, Zeyu and Shen, Haiying and Vargaftik, Shay and Basat, Ran Ben and Mitzenmacher, Michael and Yu, Minlan},
  booktitle={Proceedings of the ACM SIGCOMM 2025 Conference},
  pages={1245--1247},
  year={2025}
}

@article{yang2025qwen3,
  title={Qwen3 technical report},
  author={Yang, An and Li, Anfeng and Yang, Baosong and Zhang, Beichen and Hui, Binyuan and Zheng, Bo and Yu, Bowen and Gao, Chang and Huang, Chengen and Lv, Chenxu and others},
  journal={arXiv preprint arXiv:2505.09388},
  year={2025}
}

@inproceedings{zellers2019hellaswag,
    title={{HellaSwag}: Can a Machine Really Finish Your Sentence?},
    author={Zellers, Rowan and Holtzman, Ari and Bisk, Yonatan and Farhadi, Ali and Choi, Yejin},
    booktitle ={Proceedings of the 57th Annual Meeting of the Association for Computational Linguistics},
    year={2019}
}

@inproceedings{Bisk2020piqa,
  author = {Yonatan Bisk and Rowan Zellers and Ronan Le Bras and Jianfeng Gao and Yejin Choi},
  title = {{PIQA}: Reasoning about Physical Commonsense in Natural Language},
  booktitle = {Thirty-Fourth AAAI Conference on Artificial Intelligence},
  year = {2020},
}

@InProceedings{ai2:winogrande,
    author = {Keisuke, Sakaguchi and Ronan, Le Bras and Chandra, Bhagavatula and Yejin, Choi},
    title = {{WinoGrande}: An Adversarial Winograd Schema Challenge at Scale},
    year={2019}
}

@article{saedi2024ss1,
  title={Ss1: Accelerating inference with fast and expressive sketch structured transform},
  author={Saedi, Kimia and Desai, Aditya and Walia, Apoorv and Lee, Jihyeong and Zhou, Keren and Shrivastava, Anshumali},
  journal={Advances in Neural Information Processing Systems},
  volume={37},
  pages={48921--48954},
  year={2024}
}

@InProceedings{pmlr-v235-egiazarian24a,
  title = 	 {Extreme Compression of Large Language Models via Additive Quantization},
  author =       {Egiazarian, Vage and Panferov, Andrei and Kuznedelev, Denis and Frantar, Elias and Babenko, Artem and Alistarh, Dan},
  booktitle = 	 {Proceedings of the 41st International Conference on Machine Learning},
  pages = 	 {12284--12303},
  year = 	 {2024},
  editor = 	 {Salakhutdinov, Ruslan and Kolter, Zico and Heller, Katherine and Weller, Adrian and Oliver, Nuria and Scarlett, Jonathan and Berkenkamp, Felix},
  volume = 	 {235},
  series = 	 {Proceedings of Machine Learning Research},
  month = 	 {21--27 Jul},
  publisher =    {PMLR}
}
